\documentclass[10pt,twocolumn,letterpaper]{article}

\usepackage{cvpr}              

\usepackage[dvipsnames]{xcolor}

\definecolor{cvprblue}{rgb}{0.21,0.49,0.74}
\usepackage[pagebackref,breaklinks,colorlinks,citecolor=cvprblue]{hyperref}
\usepackage{booktabs} 
\usepackage{multirow}
\usepackage{bm}
\usepackage{fontawesome5}
\usepackage{xcolor} 
\usepackage{stfloats}
\definecolor{mesh_color}{HTML}{0066CC}
\definecolor{tex_color}{HTML}{E3A163}
\definecolor{ngp_color}{HTML}{4D9900}
\def\paperID{156} 
\def\confName{3DV\xspace}
\def\confYear{2024\xspace}

\title{MixRT: Mixed Neural Representations For Real-Time NeRF Rendering}

\author{Chaojian Li \thanks{Work done while interning at Meta. $^\dagger$Corresponding Author.}\\
Georgia Tech\\
{\tt\small cli851@gatech.edu}
\and
Bichen Wu $^\dagger$\\
Gen AI, Meta\\
{\tt\small wbc@meta.com}
\and
Peter Vajda\\
Gen AI, Meta\\
{\tt\small vajdap@meta.com}
\and
Yingyan (Celine) Lin\\
Georgia Tech\\
{\tt\small celine.lin@gatech.edu}
\and
Project Page: \url{https://licj15.github.io/MixRT}
}

\begin{document}
\maketitle
\begin{abstract}
Neural Radiance Field (NeRF) has emerged as a leading technique for novel view synthesis, owing to its impressive photorealistic reconstruction and rendering capability. Nevertheless, achieving real-time NeRF rendering in large-scale scenes has presented challenges, often leading to the adoption of either intricate baked mesh representations with a substantial number of triangles or resource-intensive ray marching in baked representations. We challenge these conventions, observing that high-quality geometry, represented by meshes with substantial triangles, is not necessary for achieving photorealistic rendering quality. Consequently, we propose MixRT, a novel NeRF representation that includes a low-quality mesh, a view-dependent displacement map, and a compressed NeRF model. This design effectively harnesses the capabilities of existing graphics hardware, thus enabling real-time NeRF rendering on edge devices. Leveraging a highly-optimized WebGL-based rendering framework, our proposed MixRT attains real-time rendering speeds on edge devices (over 30 FPS at a resolution of 1280 × 720 on a MacBook M1 Pro laptop), better rendering quality (0.2 PSNR higher in indoor scenes of the Unbounded-360 datasets), and a smaller storage size (less than 80\% compared to state-of-the-art methods).
\end{abstract}    
\begin{figure}[t]
  \centering
  \includegraphics[width=1\linewidth]{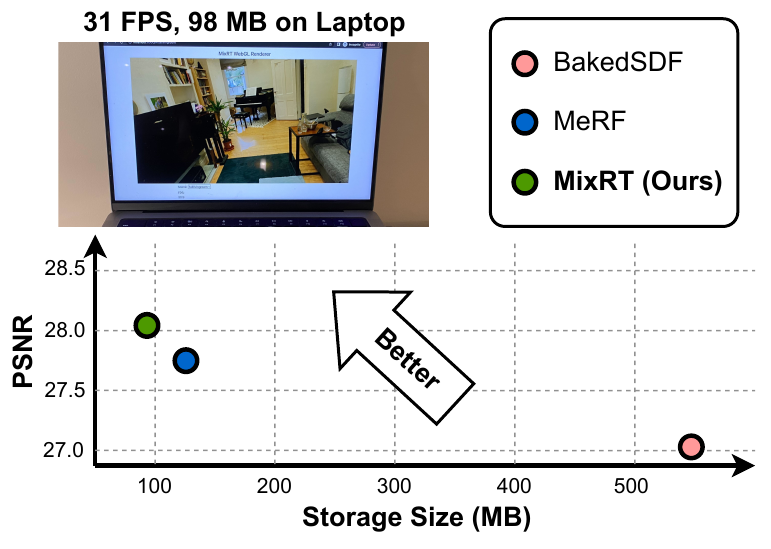}
  \vspace{-1.5em}
\caption{Our proposed MixRT can enable real-time rendering ($>$ 30 FPS) at a resolution of 1280 $\times$ 720 on a Macbook M1 Pro laptop with better rendering quality and smaller storage size compared to SotA works on real-time NeRF rendering~\cite{yariv2023bakedsdf,reiser2023merf}.}
\label{fig:teaser}
\end{figure}

\section{Introduction}
\label{sec:intro}
Neural Radiance Field (NeRF), first introduced by~\cite{mildenhall2020nerf}, has been established as the state-of-the-art (SotA) technique in novel view synthesis tasks, owing to its superior ability to deliver photorealistic rendering quality. Despite its remarkable capabilities, the practical application of NeRF, especially in immersive interactions on edge devices, has been significantly hampered due to its slow rendering speed.
Recognizing this limitation, several prior works have proposed various methods to enhance the efficiency of NeRF. These methods, such as baking NeRF into more efficient representations like mesh~\cite{chen2023mobilenerf} or sparse voxels~\cite{hedman2021baking}, have achieved impressive results, demonstrating real-time rendering speed (greater than 30 FPS) on edge devices. Unfortunately, these methods often fall short when applied to larger-scale real-world scenes, either yielding unacceptably slow rendering speeds or requiring prohibitive storage resources.
Efforts to overcome these challenges have typically focused on baking NeRF into a high-quality geometry representations (for instance, more than 10 million triangles~\cite{yariv2023bakedsdf}) or resorting to computationally costly ray marching in the baked representations~\cite{reiser2023merf}. Despite offering partial solutions to the challenges, these approaches still suffer from inherent drawbacks related to efficiency and resource requirements.

Upon careful examination, we observe a critical insight that differs from the established conventions. We identify that a high-quality mesh is not necessary in the baked representations concerning rendering quality. Our observation suggests that it is feasible to trade-off the complexity of the baked mesh for a more refined representation of color fields (such as NeRF), thereby achieving a more favorable balance between rendering quality and efficiency.
Guided by these observations, we propose MixRT, a unique NeRF representation, that mixes different neural representations for real-time NeRF rendering. Specifically, the propsoed MixRT consists of (1) a low-quality mesh (approximately 15 MB compared to over 300 MB in BakedSDF~\cite{yariv2023bakedsdf}) that provides coarse geometric information of the scene, (2) a view-dependent displacement map to calibrate the ray-mesh intersection points before fetching the corresponding color, and (3) a compressed NeRF model, in the format of an Instant-NGP~\cite{muller2022instant} that provides the density and color of each sampled point. This innovative combination not only ensures the preservation of rendering quality but also maximizes the efficient utilization of available hardware resources, including Rasterizer, Texture Mapping Units, and Single Instruction Multiple Data (SIMD) Units. This balance of resource allocation enables us to achieve real-time rendering speeds on edge devices while minimizing storage requirements, making it an ideal solution for performance-conscious applications.

In summary, our key contributions are as follows:
\begin{itemize}
    \item Through our observations, we have discovered that achieving high rendering quality in novel view synthesis tasks does not require high-complexity geometry represented by meshes with a vast number of triangles. This revelation has sparked the concept of simplifying the baked mesh and combining various neural representations. As a result, we have experienced substantial improvements in both efficiency and flexibility, paving the way for more efficient and versatile rendering techniques.
    \item We introduce an innovative NeRF representation, which consists of three essential components: a low-quality mesh, a view-dependent displacement map, and a compressed NeRF model. This carefully crafted design is specifically optimized to fully harness the capabilities of rasterizers, texture mapping units, and SIMD units in current graphics hardware. As a result, it empowers us to achieve real-time NeRF rendering on edge devices without compromising on rendering quality. 
    \item In addition, we develop a highly optimized WebGL-based rendering framework, which allows our proposed MixRT to achieve SotA rendering quality (e.g., PSNR) vs. efficiency (e.g., FPS and storage size) trade-offs.
\end{itemize}

\section{Related Works}
\subsection{NeRF on Large-Scale Scenes}
\label{sec:related_works_large_scale_nerf}
Previous works on NeRF rendering for large-scale real-world scenes can be divided into two main categories. 
Works in the first category divide the entire space into multiple sub-spaces, assigning individual NeRFs with specific radii to each. Specifically,~\cite{tancik2022block,turki2022mega} align the training of NeRFs for different sub-spaces with collected images in varying lighting conditions;~\cite{xiangli2022bungeenerf} takes this a step further, using different NeRFs for views at varying scales, allowing city-scale scene rendering;~\cite{zhenxing2022switch,fang2023nerfuser} enhance the rendering quality by selecting or fusing outputs from multiple sub-space NeRF models. 
Works in the second category map the entire space into a specific bounded space. In particular,~\cite{barron2022mip} introduces the concept of using a contraction function to fold the unbounded scene domain into a finite sphere;~\cite{reiser2023merf} later refines this function, making it piecewise for efficient computation of ray-AABB intersections;~\cite{barron2023zip,tancik2023nerfstudio} subsequently improve the contraction function further to better handle multisample isotropic Gaussian and voxel representations, respectively. 

In our approach, MixRT, we employ the contraction function outlined in~\cite{barron2022mip} to configure the NeRF model for the mapped finite sphere. Meanwhile, we retain the low-quality mesh and view-dependent displacement map in their original space. This allows us to capitalize on the optimized rasterization pipeline which is common to most graphics hardware.

\subsection{Real-Time NeRF Rendering}
\label{sec:related_works_real_time_nerf}
Real-time rendering or view synthesis is a vital and challenging problem in computer vision and graphics, given its significance in immersive interaction applications~\cite{akenine2019real}. Early techniques for real-time rendering are either dependent on a vast number of images from densely sampled viewpoints or compromised on rendering quality due to the lack of fine-grained geometry proxies during reconstruction. 
For instance,~\cite{gortler1996lumigraph,levoy1996light,mildenhall2019local} exploit light fields to interpolate target images from densely sampled images directly, while \cite{furukawa2015multi,schoenberger2016sfm,schoenberger2016mvs} utilize multi-view stereo and structure-from-motion pipelines to construct triangle meshes for real-time rendering. 
NeRF\cite{mildenhall2020nerf}, on the other hand, employs a continuous volumetric field, represented in a multi-layer perceptron (MLP) network format for scene reconstruction, achieving state-of-the-art rendering quality thanks to the ease of optimizing MLP representations through gradient descent.

Following NeRF's trailblazing results, subsequent works have proposed "baking" (i.e., pre-computing intermediate results and storing them in buffers) NeRF models into more efficient representations to achieve NeRF's high-quality rendering with real-time speeds. These efficient representations are well-optimized on existing graphics hardware and include triangle meshes or sparse voxels. 

In particular,~\cite{hedman2021baking,yu2021plenoctrees,wu2022scalable,reiser2023merf} bake NeRF models into sparse voxels with compact storage formats, enabling real-time rendering speeds with existing CUDA or WebGL APIs. On the other hand, ~\cite{chen2023mobilenerf,yariv2023bakedsdf,rakotosaona2023nerfmeshing,munkberg2022extracting} adopt triangle meshes in the rendering pipeline, distilling them from pre-trained NeRF models or training them from scratch with differentiable rendering frameworks. Moreover,~\cite{cao2023real,li2021neulf} have developed either fully convolution-based or MLP-based networks to reconstruct light fields and enable real-time NeRF rendering on mobile devices. However, their wider application is limited either by platform-dedicated deployment tools~\cite{coreml} or is constrained to synthesizing front views only. There also exist point-cloud-based works like~\cite{kerbl20233d} that utilize point clouds as scene representations for faster rendering speeds. However, their approaches, heavily relying on custom CUDA kernels for computational efficiency, face limitations in terms of broader adaptability. Specifically, the lack of compatibility with downstream computer graphics toolchains (e.g., editing and making collision animation in Blender~\cite{blender}) limits their utility across a diverse range of edge devices

Our proposed MixRT is unique among the mentioned real-time NeRF rendering methods, combining a low-quality mesh, a view-dependent displacement map, and a compressed NeRF model in the Instant-NGP format~\cite{muller2022instant}. By leveraging the rasterizers, texture mapping units, and SIMD units accessible by WebGL APIs on most existing devices, MixRT can achieve SotA rendering quality with real-time rendering speeds, suitable storage size, and memory requirements for large-scale real-world scenes (e.g., Unbounded-360 dataset~\cite{barron2022mip}). Specifically, thanks to the adoption of a rasterization-based rendering pipeline, our proposed MixRT not only supports multiple devices through a cross-platform graphics library but is also compatible with existing computer graphics toolchains (e.g., collision detection from ~\cite{cannonjs})\footnote{The real-time online collision demonstration of our proposed MixRT is available at \url{https://licj15.github.io/MixRT/collision_viewer/}}.

\section{Preliminaries}
\subsection{NeRF Rendering Pipeline}
\label{sec:preliminaries_nerf_pipeline}
NeRF~\cite{mildenhall2020nerf} offers photorealistic novel views by encoding a continuous volumetric field of points, which intercept and emit light rays, within the parameters of an MLP network. The rendering process with NeRF involves three steps. (1) To render each pixel in the target novel view, a ray $\mathbf{r} = \mathbf{o}+t\mathbf{d}$ is cast from the origin (such as the camera's center) of the target novel view $\mathbf{o}$ along direction $\mathbf{d}$, which passes through the respective pixel. Here, $t$ denotes the distance between sampled points along this ray and the origin $\mathbf{o}$. (2) For each point distanced $t_k$ from the view origin $\mathbf{o}$, its location $\mathbf{o}+t_k\mathbf{d}$ and direction $\mathbf{d}$ serve as inputs to the MLP network $(\mathbf{o}+t_k\mathbf{d}, \mathbf{d}) \rightarrow (\sigma_k, \mathbf{c}_k)$, which then outputs the corresponding density $\sigma_k$ and an RGB color $\mathbf{c}_k$. These represent the extracted features of that specific point. (3) Adhering to the principles of classical volume rendering~\cite{max1995optical}, the color $\mathbf{C}(\mathbf{r})$ of the pixel corresponding to the ray $\mathbf{r}$ can be computed by integrating the features of the points along the ray. The following equation expresses this process:

\begin{align}
\mathbf{C}(\mathbf{r}) = \sum_{k=1}^NT_k(1-\exp(-\sigma_k (t_{k+1}-t_{k})))\mathbf{c}_k, \nonumber\\
\vspace{-3em}
\textrm{ where } T_k = \exp(-\sum_{j=1}^{k} \sigma_{j} (t_{j+1}-t_{j})),
\label{eq:nerf_render}
\end{align}
where  $N$ denotes the number of sampled points along the ray $\mathbf{r}$ and $T_k$ indicates the accumulated transmittance along the ray $\mathbf{r}$ to the point $\mathbf{o}+t_k\mathbf{d}$. This transmittance represents the likelihood of the ray reaching this point without encountering any other points.

To further accelerate NeRF's reconstruction process, Instant-NGP~\cite{muller2022instant} replaces the MLP network of vanilla NeRF~\cite{mildenhall2020nerf} with a 3D embedding grid stored as a compact 1D hash table. As a result, the computationally heavy MLP inferences in the standard NeRF, involving about 1 million FLOPs, are transformed into significantly less demanding embedding interpolation operations, requiring fewer than 0.00005 million FLOPs. 
Specifically, for each queried point along the rays passing through the pixels of training images, the embeddings of its eight nearest vertices in the 3D embedding grid are retrieved from the compact 1D hash table using their respective table index that is determined by their coordinates. The embeddings of the queried point are then obtained through trilinear interpolation of these eight embeddings. 
After retrieving the embeddings for the queried points along the rays passing through the pixels as described above, these embeddings are fed into a smaller MLP model to obtain the corresponding density and view-dependent color.

Unlike the vanilla NeRF which employs an MLP with 10 layers, each with 256 hidden units, this smaller MLP comprises only 2 layers with 64 hidden units each~\cite{muller2022instant}. As discussed in Sec.~\ref{sec:preliminaries_representations} and recent real-time NeRF rendering studies~\cite{yariv2023bakedsdf,reiser2023merf}, Instant-NGP\cite{muller2022instant} retains the storage efficiency of vanilla NeRF due to the compact 1D hash table but can only achieve real-time rendering speeds on high-end GPUs, such as RTX 3090Ti~\cite{rtx3090}. The challenge remains to enable real-time rendering of large-scale real-world scenes using Instant-NGP while maintaining storage efficiency. As we analyze in Sec.~\ref{sec:method_observations}, directly combining low-quality meshes with Instant-NGP retains storage efficiency, but fails to achieve real-time rendering speeds. Informed by the profiling in Sec.~\ref{sec:method_profiling}, our proposed MixRT modifies the model structure of Instant-NGP to better align with the WebGL framework, making it more accessible for most existing devices equipped with browsers.

\subsection{Discussion on Existing NeRF Representations}
\label{sec:preliminaries_representations}

\begin{table*}[!t]
\caption{Overview of Commonly-Used NeRF Representations}
\centering
  \resizebox{\linewidth}{!}
  {
    \begin{tabular}{c||c|c|c|c|c}
    \toprule
    \textbf{Representations} & \textbf{Rendering Quality} & \textbf{FPS} & \textbf{VRAM Efficiency} & \textbf{Storage Efficiency}  & \textbf{Hardware} \\
    \midrule
    MLP Network~\cite{mildenhall2020nerf,tancik2022block} & High & Low &  High & High & SIMD Units \\
    Triangle Mesh~\cite{chen2023mobilenerf,yariv2023bakedsdf,rakotosaona2023nerfmeshing} & Medium & High & Low & Low & Rasterizer\\
    Sparse Voxels~\cite{hedman2021baking,yu2021plenoctrees} & Medium & Medium & Low & Low & Texture/SIMD Units \\
    Plane/Vector~\cite{reiser2023merf,chen2022tensorf} & Medium & Medium & Medium & Medium & Texture/SIMD Units \\
    Hash Table~\cite{muller2022instant,barron2023zip} & High & Low & High & High & SIMD Units \\
    \bottomrule
    \end{tabular}
    }
  \label{tab:representation_summary}
\end{table*}

As detailed in Sec.\ref{sec:related_works_real_time_nerf}, prior works have explored the adoption of alternative, more efficient NeRF representations in lieu of MLP networks for the purpose of real-time rendering. Yet, to date, no NeRF representation has been able to simultaneously meet the criteria of delivering high rendering quality (e.g., measured by PSNR), ensuring real-time frame rates, optimizing VRAM efficiency (which translates to minimized memory allocation during the rendering process), and maximizing storage efficiency (implying a compact model size that's conducive for efficient data transmission between users). This collective performance assessment is comprehensively summarized in Tab.~\ref{tab:representation_summary}.

Specifically, the \textbf{MLP network} used in the vanilla NeRF model~\cite{mildenhall2020nerf} excels in photorealistic rendering quality. Furthermore, it is highly efficient in terms of storage and memory usage, requiring only about 5 MB of network weights for each scene in the NeRF-Synthetic dataset~\cite{mildenhall2020nerf}. As such, it is a popular choice in subsequent research focusing on high-fidelity, large-scale NeRF rendering~\cite{tancik2022block,barron2022mip}. However, it has a significant limitation: there are no well-optimized accelerators available in the current graphics hardware to run this type of network efficiently (i.e., only SIMD units such as CUDA cores can execute the model). This results in slower rendering speeds, which restricts its application in scenarios requiring real-time interactions.

Driven by the fact that most existing edge devices support \textbf{triangle mesh} effectively within their hardware rasterizer, studies such as~\cite{chen2023mobilenerf,yariv2023bakedsdf,rakotosaona2023nerfmeshing} construct their rendering pipelines based on the mesh rasterization process. Utilizing triangle mesh as NeRF representations significantly improves FPS, enabling real-time rendering speeds even on mobile devices~\cite{chen2023mobilenerf}, while maintaining a respectable rendering quality (i.e., only 0.1 lower PSNR than that of vanilla NeRF on the NeRF-Synthetic dataset~\cite{chen2023mobilenerf}). Nevertheless, the approach's scalability remains a concern for large-scale real-world scenes, as the requirements for storage and memory increase proportionally with the scale of the scene, leading to over 400 MB of disk usage on the Unbounded-360 dataset~\cite{yariv2023bakedsdf}.

In pursuit of a better balance between the MLP network with costly volumetric ray casting and the triangle mesh with efficient rasterization, prior works like~\cite{hedman2021baking,yu2021plenoctrees} have proposed replacing the MLP network with \textbf{sparse voxels}, while still employing volumetric ray casting for the rendering process. Leveraging the compressed format of sparse voxels (for instance, densely packed 3D texture in~\cite{hedman2021baking}) and the same volumetric ray casting technique used by vanilla NeRF, these works achieve a respectable compromise between rendering quality and FPS. They utilize either the texture mapping units accessible via WebGL API or the SIMD units accessible through CUDA APIs. However, akin to the triangle mesh representations, scaling these methods to large-scale scenes can pose significant challenges in terms of memory and storage efficiency, as pointed out by~\cite{reiser2023merf}. 

In the effort to enhance the memory and storage efficiency of sparse voxels, studies such as~\cite{reiser2023merf,chen2022tensorf} propose using \textbf{plane/vector} as NeRF representations, which can be perceived as the low-rank decomposed format of 3D voxels. By employing a similar rendering pipeline and hardware as used by sparse voxels, yet with a more compact representation alongside a distinct decoding method (for instance, matrix-vector outer product in~\cite{chen2022tensorf}) for embeddings of the sampled points, these studies manage to maintain similar or even superior rendering quality vs. FPS trade-offs when compared to those using sparse voxels. In addition, they significantly reduce the memory and storage requirements (e.g., only requiring 188 MB vs. 3785 MB of disk space as per~\cite{reiser2023merf}).

Following~\cite{muller2022instant} to employ a \textbf{hash table} as a new NeRF representation, numerous subsequent studies have sought to enhance its balance between rendering quality and efficiency, given its state-of-the-art training speed. As validated by~\cite{muller2022instant,reiser2023merf}, hash tables exhibit superior memory and storage efficiency compared to sparse voxels or plane/vector (for instance, only requiring $\sim$100 MB vs. $\sim$200 MB or $\sim$400 MB of disk space in the Unbounded-360 dataset~\cite{muller2022instant,reiser2023merf}). Consequently, they are often used as the NeRF representation during training, before being translated into other representations~\cite{muller2022instant,reiser2023merf}. Additionally,\cite{barron2023zip} also affirms that a hash table representation with a well-designed point sampling strategy in ray casting can yield superior rendering quality compared to MLP network representations. Nonetheless, the hash table proposed in Instant-NGP\cite{muller2022instant} is constrained by its limited compatibility with most devices, thus only achieving real-time rendering speeds on high-end GPUs with customized CUDA kernel for accessing SIMD units in graphic hardware.

Motivated by the aforementioned comparison, discussion, and analysis of previous works, we propose a new form of NeRF representation. Our proposed MixRT mixes a low-quality mesh, a view-dependent displacement map, and a compressed NeRF model in Instant-NGP's hash table~\cite{muller2022instant} format. This configuration is purposefully designed to leverage the inherent strengths of rasterizers, texture mapping units, and SIMD units in current graphics hardware. The proposed method empowers real-time NeRF rendering on edge devices while maintaining better rendering quality (e.g., 0.2 PSNR higher on indoor scenes of the Unbounded-360 dataset), and staying within smaller memory and storage parameters (e.g., consuming only 80\% of ~\cite{reiser2023merf}'s disk usage), as compared to SotA methods.

\begin{figure*}[t]
  \centering
  \includegraphics[width=1\linewidth]{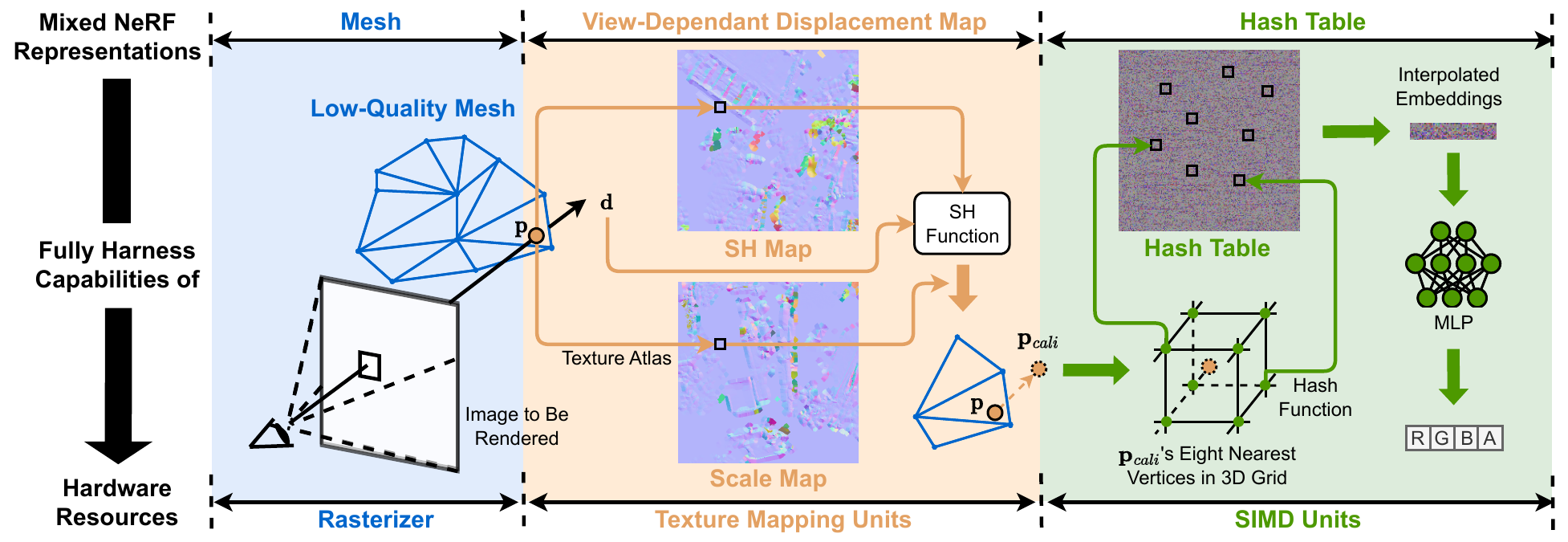}
  \vspace{-1.5em}
\caption{An overview of our proposed MixRT rendering pipeline: MixRT integrates three core components: a low-quality \textcolor{mesh_color}{\textbf{mesh}}, a \textcolor{tex_color}{\textbf{view-dependent displacement map}}, and a NeRF model compressed into a \textcolor{ngp_color}{\textbf{hash table}}. This combination aims to maximize utilization of diverse hardware resources. To render an image pixel: (1) We use \textcolor{mesh_color}{\textbf{rasterizer}} hardware to perform mesh rasterization, determining the ray-mesh intersection point, $\mathbf{p}$. (2) Leveraging \textcolor{tex_color}{\textbf{texture mapping units}}, we use texture coordinates to access maps containing the spherical harmonics (SH) coefficients and scale, computing the calibrated point, $\mathbf{p}_{cali}$. (3) Lastly, $\mathbf{p}_{cali}$ is processed by \textcolor{ngp_color}{\textbf{SIMD units}}, retrieving embeddings for its eight closest vertices from the 3D grid stored as a hash table. A small MLP network then converts these interpolated embeddings into the final rendered color.}
\label{fig:pipeline_overview}
\end{figure*}

\section{Method}
In this section, we first examine the relationship between mesh quality and rendering quality in Sec.\ref{sec:method_observations}, finding that, with the help of color fields that are represented by hash tables, high-quality novel view synthesis doesn't necessarily demand meshes with extensive triangles. We then perform the runtime profiling analysis on hash tables in Sec.\ref{sec:method_profiling}, pinpointing bottlenecks to inform hash table configuration adjustments for improved FPS. Finally, we unveil MixRT, with the detailed design describled in Sec.~\ref{sec:method_mix}, comprising: (1) a low-quality mesh, (2) a view-dependent displacement map, and (3) an compressed NeRF model stored in a hash table. This design ensures MixRT's rendering pipeline, illustrated in Fig.~\ref{fig:pipeline_overview}, can be specifically optimized to fully harness the capabilities of rasterizers, texture mapping units, and SIMD units in current graphics hardware, enabling real-time NeRF rendering on edge devices without rendering quality sacrifice.

\begin{table*}[!b]
\caption{Comparison between (1) the \textbf{high-quality triangle mesh} from~\cite{yariv2023bakedsdf} and (2) the combination of  the \textbf{low-quality mesh} (simplified from the high-quality mesh) and Instant-NGP's~\cite{muller2022instant} \textbf{hash table}, in terms of the average PSNR vs. storage size or FPS trade-offs on the indoor scenes of Unbounded-360 dataset. The FPS was measured on a Macbook M1 Pro laptop with a resolution of 1280 $\times$ 720.}
\centering
  \resizebox{\linewidth}{!}
  {
    \begin{tabular}{c||c|c|c|c|c|c|c|c|c|c|c}
    \toprule
    \multirow{2}{*}{\textbf{Representations}} & \multicolumn{4}{c|}{\textbf{$\downarrow$ \# of Vertices on}}  & \multicolumn{4}{c|}{\textbf{$\downarrow$ \# of Faces on}} & \multirow{2}{*}{\textbf{$\uparrow$ Avg. PSNR}} & \multirow{2}{*}{\textbf{$\downarrow$ Storage}} & \multirow{2}{*}{\textbf{$\uparrow$FPS}}\\
     & Room & Counter & Kitchen & Bonsai & Room & Counter & Kitchen & Bonsai & & \\
    \midrule
    Mesh from~\cite{yariv2023bakedsdf} & 7,060,849 & 11,950,574 & 13,539,203 & 13,343,679 & 14,110,659 & 23,892,064 & 27,056,127 & 26,679,898 & 27.06 & 542 MB & 120 \\
    Simplified Mesh + Hash Table & 946,962 & 1,572,959 & 1,778,283 & 1,750,341 & 1,893,695 & 3,147,635 & 3,557,514 & 3,501,683 & \textbf{27.41} & \textbf{139 MB} & 0.4  \\
    \bottomrule
    \end{tabular}
    }
  \label{tab:observation}
\end{table*}

\begin{table}[!t]
\caption{Adjusting the model structure of Instant-NGP's hash table~\cite{muller2022instant}. FPS was measured on a Macbook M1 Pro laptop at a resolution of 1280 $\times$ 720, and the fragment shader was set to query the hash table for color once per pixel. ``Hash table size'' and ``\# of levels'' denote the maximum entries per level and the number of multi-resolution levels in Instant-NGP's hash table, respectively. The ``MLP architecture'' outlines the structure of the MLP responsible for transforming the embedding retrieved from the hash table into RGB color.}
\centering
  \resizebox{\linewidth}{!}
  {
    \begin{tabular}{c|c|c||c}
    \toprule
    \textbf{\# of Levels} & \textbf{Hash Table Size} & \textbf{MLP Architecture} & \textbf{FPS}  \\
    \midrule
    8 & $2^{17}$ & 2 Layers, 8 Hidden Neurons & 27 \\
    8 & $2^{17}$ & \textbf{Removed} & 30 \\
    \multicolumn{4}{c}{\faHandPointRight[regular] Observation: \textbf{MLP} is not the runtime bottleneck} \\
    \midrule
    8 & $2^{17}$ & Removed & 30 \\
    \textbf{1} & $2^{17}$ & Removed & 120 \\
    \multicolumn{4}{c}{\faHandPointRight[regular] Observation: \textbf{\# of levels} is the runtime bottleneck} \\
    \midrule
    8 & $2^{17}$ & Removed & 30 \\
    8 & \bm{$2^{5}$} & Removed & 35 \\
    8 & \bm{$2^{22}$} & Removed & 25 \\
    \multicolumn{4}{c}{\faHandPointRight[regular] Observation: \textbf{Hash table size} is not the bottleneck} \\
    \bottomrule
    \end{tabular}
    }
  \label{tab:profiling}
\end{table}

\subsection{Observations on the Effect of Mesh Quality}
\label{sec:method_observations}

Previous real-time NeRF rendering research that uses triangle mesh as the NeRF representation highlights the importance of mesh geometry quality for photorealistic rendering~\cite{yariv2023bakedsdf,tang2023delicate}. For instance,~\cite{tang2023delicate} delves into refining the surface by adjusting vertex positions and face density. Yet, we note that an ultra-detailed mesh, packed with a vast number of triangles, isn't mandatory for photorealistic rendering outcomes.

In particular, we made the above observations by (1) simplifying the SotA high-quality mesh from~\cite{yariv2023bakedsdf} via the classical vertex clustering~\cite{open3dvertexclustering} and (2) querying the color of the ray-mesh intersection points from a color field represented by Instant-NGP~\cite{muller2022instant}'s hash table. As summarized in Tab.~\ref{tab:observation}, the low-quality mesh contains more than 5 $\times$ fewer triangles and faces as compared to the high-quality one but can achieve $\sim$ 0.3 higher PSNR than the high-quality mesh by equipping a hash table as a color filed to query the color from. Thanks to the high memory and storage efficiency of Instant-NGP~\cite{muller2022instant}'s hash table, the combination described above only consumes 0.26 $\times$ storage size of the high-quality mesh. However, as suggested in prior works~\cite{reiser2023merf,yariv2023bakedsdf} and the discussion in Sec.~\ref{sec:preliminaries_representations}, the achieved FPS by the combination of simplified mesh and hash table can not satisfy the real-time rendering requirements. The set of conducted experiments implies that (1) the high-quality geometry information represented by mesh with massive triangles is not necessary for achieving high rendering quality and (2) replacing high-quality mesh with the combination of the simplified low-quality mesh and a hash table as the NeRF representation can achieve better PSNR vs. storage efficiency trade-offs.

From the above observations, it is evident that the primary limitation of merging a low-quality mesh with other representations is the rendering speed. Therefore, we conduct an in-depth runtime profiling analysis on the hash table representation which is known for its memory and storage efficiency, as detailed in Sec.~\ref{sec:method_profiling}.

\begin{table}[!b]
\caption{Optimizing Instant-NGP's hash table configurations based on runtime profiling insights. FPS measurements were taken on a Macbook M1 Pro laptop at a resolution of 1280 $\times$ 720, while PSNR evaluations were conducted on the indoor scenes of the Unbounded-360 dataset.}
\centering
  \resizebox{\linewidth}{!}
  {
    \begin{tabular}{c|c|c||c|c|c}
    \toprule
    \textbf{Mesh} & \textbf{\# of Levels} & \textbf{Hash Table Size} & \textbf{$\uparrow$Avg. PSNR} & \textbf{$\downarrow$Storage} & \textbf{$\uparrow$ FPS}\\
    \midrule
    ~\cite{yariv2023bakedsdf} & - & - & 27.06 & 542 MB & 120 \\
    Simplified & 16 & $2^{20}$ & 27.41 & 139 MB & 0.4 \\
    Simplified & 4 & $2^{21}$ & 26.63 & \textbf{74 MB} & \textbf{35} \\
    \bottomrule
    \end{tabular}
    }
  \label{tab:tuning}
\end{table}

\subsection{Runtime Profiling Analysis}
\label{sec:method_profiling}

Since there is no existing runtime breakdown analysis tool for WebGL, we perform the runtime profiling analysis on hash tables by tuning the model structure and observing the resulting FPS. Specifically, as summarized in Tab.~\ref{tab:profiling}, the effect of varying (1) hash table size, (2) number of levels, and (3) MLP architectures on FPS implies that the number of levels is the runtime bottleneck, i.e., tuning it can significantly change the resulting FPS, while the other two factors are not. 

Motivated by the profiling analysis above, we propose to modify the default model structure of Instant-NGP's hash table by shrinking the number of levels while enlarging the hash table size. As suggested in Tab.~\ref{tab:tuning}, such modifications can boost the rendering speed to $>$ 30 FPS while maintaining hash tables' memory or storage efficiency.

\subsection{Mixing Mesh, Texture Map, and NeRF}
\label{sec:method_mix}

With the observation that high-quality mesh is not necessary for achieving high rendering quality (see Sec.~\ref{sec:method_observations}) and the profiling-inspired hash table configuration can achieve both high rendering speed and high memory or storage efficiency (see Sec.~\ref{sec:method_profiling}), we propose a type of NeRF representation that comprises (1) a low-quality mesh, (2) a view-dependant displacement map, and (3) a compressed NeRF model in Instant-NGP~\cite{muller2022instant}'s hash table format. Such a design can not only leverage the commonly agreed high rendering quality, high renderings speed, and high memory and storage efficiency of existing NeRF representations as discussed above but also fully leverage the rasterizers, texture mapping units, and SIMD units in graphics hardware. We summarize the rendering pipeline of our proposed MixRT in Fig.~\ref{fig:pipeline_overview} and detail the design of each part in our proposed MixRT as follows.

\subsubsection{Triangle Mesh}
We leverage the standard triangle mesh format to include the information on geometric vertices coordinates, texture coordinates, and polygonal face elements. Unlike~\cite{yariv2023bakedsdf}, we do not need to store the per-vertex appearance parameters because the color of the intersection points will be fetched from the hash table. Following~\cite{yariv2023bakedsdf}, the mesh is post-processed by vertex order optimization~\cite{sander2007fast} to allow higher cache hit rates for accessing neighboring triangles. 

\subsubsection{View-Dependant Displacement Map}
Inspired by the commonly-used normal map~\cite{cohen1998appearance,cignoni1998general} that fakes the lighting of bumps and dents without using more polygons, we propose a view-dependant displacement map to calibrate the coordinate of the intersection points to be inputted in the color field represented by Instant-NGP's hash table. Similar to a normal map, our proposed view-dependant displacement map can fake more accurate coordinates of the intersection points without adding new polygons to the mesh. However, our proposed one can better fit Instant-NGP~\cite{muller2022instant}'s hash table and the corresponding rendering pipeline that only takes the coordinates and view directions as input instead of surface normal. While previous studies~\cite{attal2023hyperreel,fang2021neusample} use neural networks to predict displacement vectors for calibrating points in volumetric rendering, this approach is unfeasible for real-time on-device rendering. In contrast, our method employs 2D maps for displacement prediction, leveraging the texture mapping units of graphics hardware.

In particular, the proposed view-dependant displacement map consists of (1) a spherical harmonics (SH) map, $m_{SH}$, to store the SH coefficients for guiding the encoded view directions to output a view-dependant vector and (2) a scale map $m_{s}$ to scale the outputted view-dependant vector to a proper length and thus the scaled vector can be used as the calibration variable of the coordinate of the ray-mesh intersection points. The shape of the SH map $m_{SH}$ and scale map $m_{s}$ is designed to be $[R_m, R_m, 3 \times (D_{SH}+1)^2]$ and  $[R_m, R_m, 1]$, respectively. $R_m$ represents the resolution of the map and $D_{SH}$ is the SH degree used in SH map $m_{SH}$. Given the coordinate $\mathbf{p} \in \mathbb{R}^3$ of an intersection point, its texture coordinates $\mathbf{p}_t \in \mathbb{R}^2$, and the corresponding view direction $\mathbf{d} \in \mathbb{R}^3$, the calibrated coordinate $\mathbf{p}_{cali} \in \mathbb{R}^3$ can be computed as:

\begin{align}
\mathbf{p}_{cali} = \mathbf{p} + S(m_{SH}(\mathbf{p}_t), \mathbf{d}) \times m_{s}(\mathbf{p}_t),
\label{eq:vdd}
\end{align}
where $S$ denotes the SH functions as used in~\cite{chen2022tensorf}. $m_{SH}(\mathbf{p}_t) \in \mathbb{R}^{(D_{SH}+1)^2}$ and $m_{s}(\mathbf{p}_t) \in \mathbb{R}$ represent the feature interpolated from $m_{SH}$ and $m_{s}$ with coordinate $\mathbf{p}_t$, respectively. As such, the calibrated coordinate $\mathbf{p}_{cali}$ can be determined by the coordinate and view directions of the ray-mesh intersection point. The view-dependant displacement map is quantized into 8 bits after training for both higher rendering speeds and memory or storage efficiency. 

\subsubsection{Hash Table}

For the hash table in the proposed MixRT, similar to the settings in Instant-NGP~\cite{muller2022instant}, it consists of (1) multiple levels of 1D hash tables with different corresponding 3D resolutions and (2) small MLP networks to convert the fetched embeddings from the hash table to density or color. However, as illustrated in Sec.~\ref{sec:method_profiling} we modify its model structure, i.e., shrinking the number of levels and enlarging the hash table size, to improve its rendering speed. To be compatible with the 2D texture mapping units that can be accessed by WebGL, we reshape the hash tables stored in 1D format to 2D image format for exporting it to the WebGL rendering framework.

\section{Experiments}
\subsection{Experiments Settings}
\subsubsection{Baselines, Datasets, and Metrics}
We benchmark the proposed MixRT on challenging large-scale indoor scenes of the Unbounded-360 dataset~\cite{barron2022mip} and compare with the following three SotA real-time NeRF rendering works: (1) BakedSDF~\cite{yariv2023bakedsdf}: it leverages high-quality mesh with massive triangles, and gets the rendering results by mesh rasterization with appearance parameters stored in each vertex, (2) NeRFMeshing~\cite{rakotosaona2023nerfmeshing}: it also leverages high-quality mesh that is distilled from pre-trained NeRF model but store the appearance parameters as texture maps, and (3) MeRF~\cite{reiser2023merf}, which adopts tri-planes as the representations to store density and color information of the scene and gets the rendering results by volumetric ray casting like vanilla NeRF~\cite{mildenhall2020nerf}. The rendering quality is measured by PSNR, and the FPS is measured on a Macbook M1 Pro laptop at the resolution of 1280 $\times$ 720, following the settings used in~\cite{reiser2023merf}. The memory or storage efficiency is measured by the total file sizes of meshes in glTF format, textures in PNG format, and scene configurations in JSON format.

\subsubsection{Implementation Details}
We implement our real-time rendering pipeline with WebGL framework. Specifically, in the GLSL vertex shader, we compute the coordinates of the ray-mesh intersection points and their corresponding texture coordinates. In the GLSL fragment shader, we first calibrate the coordinate of the intersection points with the texture coordinates and view directions, following Eq.~\ref{eq:vdd}. Then, following the standard pipeline of Instant-NGP~\cite{muller2022instant}, we loop over the hash tables' number of levels to get the trilinear interpolated embeddings from each level. The interpolated embeddings from different levels are concatenated together to be the input of a small MLP model that is implemented as matrix-vector multiplication, following the implementation in~\cite{chen2023mobilenerf}. In all our experiments, the hash table is configured to have four levels with a minimum level resolution of 256 and a maximum of 4096, and the hash map size is set as ${2^{21}}$, with each entry holding a four-dimensional vector. For the view-dependant displacement map, the map resolution $R_m$ is set as 1536 for all scenes. For the low-quality meshes, they are simplified from the SotA BakedSDF~\cite{yariv2023bakedsdf}'s mesh by the classical vertex cluster~\cite{open3dvertexclustering} algorithm in the space contracted by Mip-NeRF-360~\cite{barron2022mip}'s contraction function with the voxel size hyperparameter set as 0.01 for all scenes. The simplified meshes, along with randomly initialized view-dependent displacement maps and hash tables, are jointly trained using the loss function based on the differences between the rendered images and the ground truth images.

\subsection{Comparing with SotA}
\label{sec:exp_sota}
\begin{table}[!t]
\caption{Comparison between our MixRT and SotA real-time NeRF rendering techniques on the four indoor scenes from the Unbounded-360 Dataset~\cite{barron2022mip}. PSNR values were measured using Mip-NeRF-360's settings~\cite{barron2022mip}, while FPS was measured on a Macbook M1 Pro at a resolution of 1280 $\times$ 720, consistent with the experiment settings in~\cite{reiser2023merf}.}
\centering
  \resizebox{\linewidth}{!}
  {
    \begin{tabular}{c||c|c|c|c|c|c|c}
    \toprule
    \multirow{2}{*}{\textbf{Method}} & \multicolumn{5}{c|}{\textbf{$\uparrow$ PSNR on}} & \multirow{2}{*}{\textbf{$\uparrow$ FPS}} & \multirow{2}{*}{\textbf{$\downarrow$ Storage}}\\
     & Room & Counter & Kitchen & Bonsai & Avg. & & \\
        \midrule
    NeRFMeshing~\cite{rakotosaona2023nerfmeshing} & 26.13 & 20.00 & 23.59 & 25.58 & 23.83 & - & - \\
    BakedSDF~\cite{yariv2023bakedsdf} & - & - & - & - & 27.06 & 120 & 542 MB\\
    MeRF~\cite{reiser2023merf} & - & - & - & - & 27.80 & 30 & 124 MB\\
    \textbf{MixRT (Ours)} & 29.88 & 26.60 & 27.46 & 28.10 & \textbf{28.01} & 31 & \textbf{98 MB} \\
    \bottomrule
    \end{tabular}
    }
  \label{tab:compare_sota}
\end{table}

We first compare our proposed MixRT with SotA real-time NeRF rendering works. As summarized in Tab.~\ref{tab:compare_sota}, the proposed MixRT achieves the highest PSNR and storage efficiency among all the methods in the benchmark, while maintaining the real-time ($>$ 30 FPS) rendering speed. Specifically, as compared to MeRF~\cite{reiser2023merf}, our proposed MixRT achieves 0.2 higher PSNR than it with only 80\% storage cost under the same rendering speed. Please refer to Appendix~\ref{sec:vis_comp} for the corresponding qualitative comparison.

\subsection{Ablation Study}
\label{sec:exp_abl}

\begin{table}[!b]
\caption{Comparison MixRT w/ and w/o the proposed view-dependent displacement (VDD) map, in terms of PSNR, FPS, and storage size on the indoor scenes of the Unbounded-360 dataset~\cite{barron2022mip}.}
\centering
  \resizebox{\linewidth}{!}
  {
    \begin{tabular}{c||c|c|c|c|c|c|c}
    \toprule
    \multirow{2}{*}{\textbf{Method}} & \multicolumn{5}{c|}{\textbf{$\uparrow$ PSNR on}} & \multirow{2}{*}{\textbf{$\uparrow$ FPS}} & \multirow{2}{*}{\textbf{$\downarrow$ Storage}}\\
     & Room & Counter & Kitchen & Bonsai & Avg. & & \\
    \midrule
    MixRT \textbf{w/} VDD Map & 29.88 & 26.60 & 27.46 & 28.10 & \textbf{28.01} & 31 & 98 MB \\
    MixRT \textbf{w/o} VDD Map & 29.10 & 25.26 & 25.64 & 26.54 & 26.64 & 35 & \textbf{74 MB} \\
    \bottomrule
    \end{tabular}
    }
  \label{tab:abl_vdd}
\end{table}

As highlighted in Sec.~\ref{sec:method_observations}, incorporating hash tables into our MixRT framework is critical for maintaining high memory and storage efficiency without sacrificing rendering quality. Following this, we further conduct an ablation study to verify the significance of the view-dependent displacement map, another integral component of MixRT. As shown in Tab.~\ref{tab:abl_vdd}, removing the view-dependent displacement map from our proposed MixRT reduces storage by approximately 24\% but results in a 1.37 decrease in PSNR. Meanwhile, the rendering speed remains relatively stable, shifting from 31 FPS to 35 FPS. Considering that MixRT already achieved higher storage efficiency than all baselines, as demonstrated in Sec.~\ref{sec:exp_sota}, integrating view-dependent displacement maps in our MixRT is a better option to achieve higher PSNR vs. rendering speed trade-offs.

\section{Limitation}

\begin{table}[!t]
\caption{Comparison between our MixRT and SotA real-time NeRF rendering techniques on the three publicly available outdoor scenes from the Unbounded-360 Dataset~\cite{barron2022mip}.}
\centering
  \resizebox{0.75\linewidth}{!}
  {
    \begin{tabular}{c||c|c|c|c}
    \toprule
    \multirow{2}{*}{\textbf{Method}} & \multicolumn{4}{c}{\textbf{$\uparrow$ PSNR on}} \\
     & Bicycle & Garden & Stump & Avg. \\
        \midrule
    \multicolumn{5}{c}{Volumetric-Rendering-Based Methods} \\
    \midrule
    MeRF~\cite{reiser2023merf} &  22.82 & 25.32 & 25.06 & 24.40 \\
    \bottomrule
    \multicolumn{5}{c}{Rasterization-Based Methods} \\
    \midrule
    MobileNeRF~\cite{chen2023mobilenerf} & 21.70 & 23.53 & 23.95 & 21.06\\
    NeRFMeshing~\cite{rakotosaona2023nerfmeshing} & 21.15 & 22.91 & 22.66 & 22.24 \\
    \textbf{MixRT (Ours)} & 21.81 & 24.55 & 23.76 & 23.37 \\
    \bottomrule
    \end{tabular}
    }
  \label{tab:outdoor}
\end{table}

As illustrated in Tab.~\ref{tab:compare_sota}, our proposed MixRT demonstrates better rendering quality vs. rendering speeds and storage efficiency than SotA methods in indoor scenes. However, its rendering quality is still constrained by rasterization-based rendering methods, a common limitation in rasterization-based real-time NeRF methods~\cite{chen2023mobilenerf,yariv2023bakedsdf}. In particular, for the more complex outdoor scenes from the Unbounded-360 dataset~\cite{barron2022mip}, as shown in Tab.~\ref{tab:outdoor}, MixRT's rendering quality is 1 PSNR lower than the volumetric-rendering-based MeRF~\cite{reiser2023merf}. Despite this, it still achieves comparable or better quality than other rasterization-based baselines~\cite{chen2023mobilenerf,rakotosaona2023nerfmeshing}.

\section{Conclusion}

We present MixRT, a NeRF representation that combines a low-quality mesh, a view-dependent displacement map, and a compressed NeRF in a hash table structure. This design emerges from our observation that achieving high rendering quality does not require high-complexity geometry represented by meshes with a vast number of triangles. This realization suggests the potential to streamline the baked mesh and incorporate diverse neural representations for rendering, memory, and storage efficiency. Through detailed runtime profiling analysis and an optimized WebGLbased rendering framework, MixRT offers state-of-the-art balance between rendering quality and efficiency.

\section*{Acknowledgement}
Chaojian Li and Yingyan (Celine) Lin would like to acknowledge the funding support from the National Science Foundation (NSF) Computing and Communication Foundations (CCF) programs (Award ID: 2211815 and 2312758).

{
    \small
    \bibliographystyle{ieeenat_fullname}
    \bibliography{main}
}
\clearpage
\setcounter{page}{1}
\maketitlesupplementary

\appendix
\section{Real-Time Interactive Demonstration}
To experience the real-time interactive demonstration of the proposed MixRT, please visit \url{https://licj15.github.io/MixRT/index.html#demos}. Our demo offers real-time online interaction with static scenes, as well as collision animations.

\section{Visual Comparison with SotA}
\label{sec:vis_comp}
In addition to the quantitative comparison of our proposed MixRT and the SotA real-time NeRF rendering shown in Tab~\ref{tab:compare_sota}, we offer additional rendered image comparisons in Fig.~\ref{fig:vis_supp} below. Consistent with the observations in Sec.~\ref{sec:exp_sota}, our proposed MixRT excels in two main areas: (1) accurately rendering regions with specular highlights or fine-grained geometry structures, e.g., the bowl in ``Scene: Counter'' and the bulldozer bucket in ``Scene: Lego'', and (2) eliminating ghostly effects such as the "floaters" observed on the floor of ``Scene: Room'' and the wall of ``Scene: Bonsai''.

\begin{figure*}[t]
  \centering
  \includegraphics[width=1\linewidth]{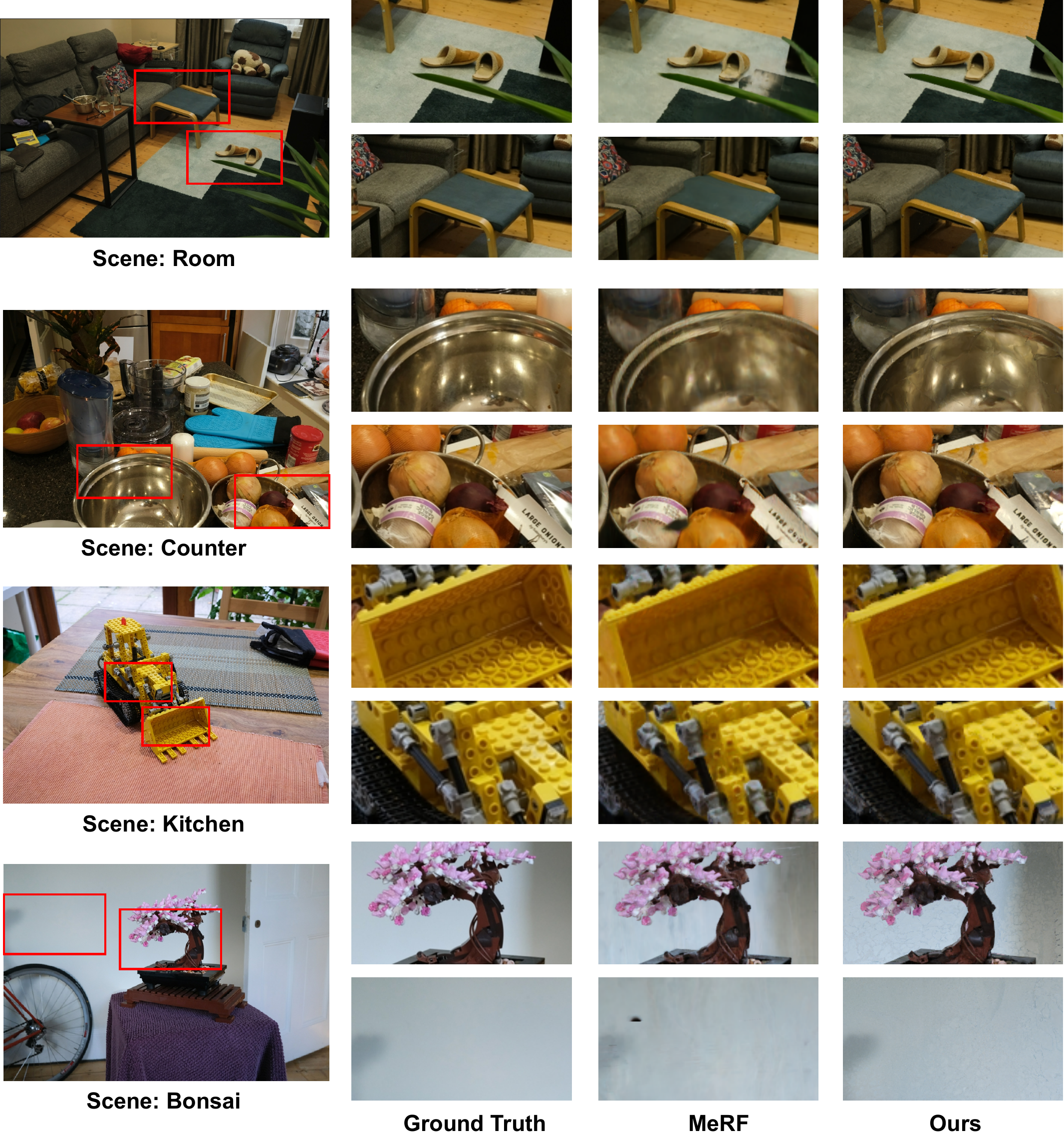}
  \vspace{-1.5em}
\caption{Visual comparison between our proposed MixRT and MeRF~\cite{reiser2023merf}, a real-time NeRF rendering work with SotA rendering quality vs. efficiency trade-offs. The rendered images are randomly selected from the test set.}
\label{fig:vis_supp}
\end{figure*}

\end{document}